\documentclass[10pt,twocolumn,letterpaper]{article}
\usepackage{booktabs}   
\usepackage{tabularx}  
\usepackage{caption}   
\usepackage[blocks]{authblk}
\usepackage{xcolor}
\usepackage{listings}

\usepackage{cvpr}              % To produce the CAMERA-READY version
\definecolor{cvprblue}{rgb}{0.21,0.49,0.74}
\usepackage[pagebackref,breaklinks,colorlinks,allcolors=cvprblue]{hyperref}

\def\paperID{CV4CHL-7} % *** Enter the Paper ID here
\def\confName{CVPR}
\def\confYear{2026}

\title{The 1st AI Children Challenge}

\author{
    Boyi Li\textsuperscript{3}, 
    Yifan Shen\textsuperscript{3}, 
    Houze Yang\textsuperscript{3}, 
    Xu Cao\textsuperscript{2,3}, 
    Guojun Yun\textsuperscript{1},
    Li Gao\textsuperscript{1}, \\\vspace{-0.3cm}
    Turong Chen\textsuperscript{1}, 
    Long Xu\textsuperscript{4}, 
    Jianguo Cao\textsuperscript{1,2}, 
    Meihuan Huang\textsuperscript{1,5,}\thanks{Corresponding author}
    \vspace{0.2cm} \\
    \textsuperscript{1}Shenzhen Children's Hospital \quad
    \textsuperscript{2}PediaMed AI \quad
    \textsuperscript{3}University of Illinois Urbana-Champaign\\
    \textsuperscript{4}Xiangya Bo'ai Rehabilitation Hospital \quad
    \textsuperscript{5}The Hong Kong Polytechnic University
    \vspace{0.1cm} \\
    {\tt\small boyil2@illinois.edu; meihuan.huang@connect.polyu.hk}
}

\begin{document}
\maketitle
\begin{abstract}
The 1st AI Children Challenge aims to advance real-world applications of computer vision and AI in child healthcare, child education, and pediatrics. The 2026 CV4CHL edition featured the first track in this domain: Children's Gait Visual Analysis. The main goal of Children's Gait Visual Analysis is the fine-grained analysis of children's gait behaviors from keypoint sequences. This is still a big challenge for human action recognition. Experienced medical doctors can distinguish these subtle nuances, but none of the people test AI models in this domain. To bridge this gap, we introduce thousands of 2D children keypoint sequences for children's walking motions across various age groups of children (3-16 years old). There is a significant opportunity for batch analysis of these sequences to provide clinically relevant insights into medical diagnosis. The Challenge will be launched with two problem tracks: \textbf{Edinburgh Visual Gait Score (EVGS) Scoring} and \textbf{Classification of Gait Patterns in Bilateral Spastic Cerebral Palsy}. Each track is chosen in consultation with board-certified pediatricians based on the value of potential solutions. With the first available dataset for such tasks and ground truth for each track, the challenge enabled participants to evaluate their solutions. Final rankings will be revealed after the competition concludes, fostering reproducibility and mitigating overfitting.

\end{abstract}

\section{Introduction}
\label{sec:intro}

\begin{figure}[!t]
    \centering
    \includegraphics[width=0.99\linewidth]{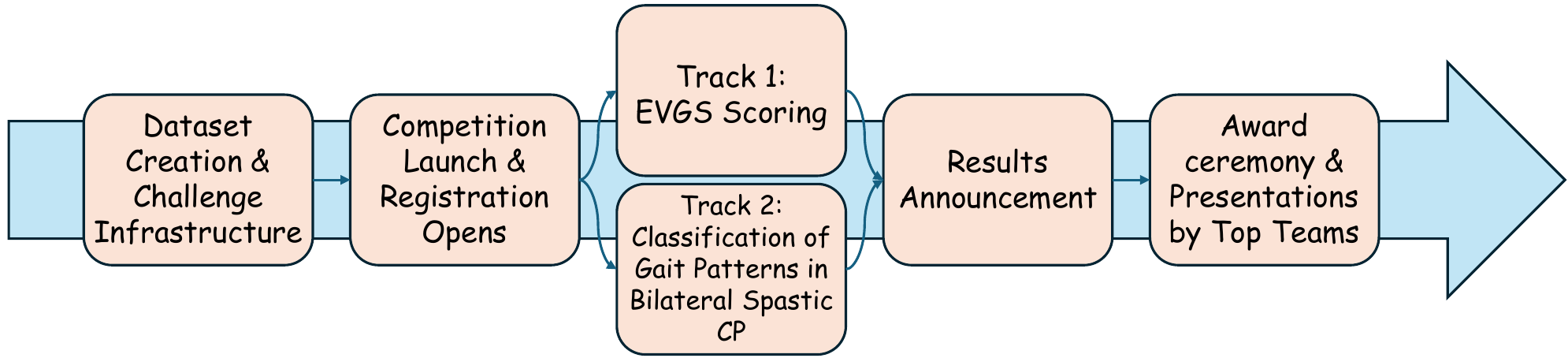}
    \caption{\textbf{The Life Cycle of the 1st AI Children Challenge.} The challenge spanned the entire process, from preparation and launch to the award ceremony. Participants may enter both Track 1 and Track 2 simultaneously.}
    \label{fig:cycle}
    \vspace{-5mm}
\end{figure}

Quantitative gait analysis is a crucial component of clinical diagnostics and physical rehabilitation for movement disorders~\cite{tan2026gaitdynamics, baker2016gait, bonanno2023gait, sharma2024factors, ben2023quantitative}.  Understanding pediatric gait patterns through various developmental milestones is crucial for diagnosing and managing several developmental disorders~\cite{armand2016gait,rathinam2014observational,baker2006gait,yun2023selective}. The early identification of gait abnormalities in conditions such as cerebral palsy primarily depends on subjective visual assessments conducted by a pediatrician during standard office visits~\cite{maathuis2005gait,dickens2006validation}. Although research using 3D gait analysis and Inertial Measurement Units (IMUs) has successfully identified quantitative warning signs for gait deviations in cerebral palsy~\cite{cook2003gait,deluca1997alterations}, implementing these technologies in clinical practice encounters significant challenges. For example, the cost of 3D gait analysis device is high and children may have uncooperative engagement. Besides, existing computer vision methods for gait analysis have largely focused on healthy adult subjects~\cite{ranjan2025computer, chen2022computer}, relying on the assumption that walking is a mature, stable, and highly rhythmic process~\cite{fan2023opengait}. Such adult-centric foundation models fail to account for the high entropy, significant intra-class variance, and inconsistent motion patterns inherent in the developing motor system~\cite{sutherland1997development,vielemeyer2026full}. \looseness=-1

To explore potential solutions to these issues in greater depth, we define a new challenge for the computer vision community: \emph{the fine-grained analysis of children's gait from video, targeting clinically-relevant characterizations of children's gait quality.}, and will host the first edition of the AI Children Challenge in which we introduce the Children Gait Visual Analysis track of this new domain, aiming to drive advancements in computer vision and AI for real-world, high-impact domains.~\Cref{fig:cycle} illustrates the life cycle of the challenge. Preparation for the Challenge began in January 2026, and the Challenge will end in the CVPR 2026 Workshop on Computer Vision for
Children (CV4CHL).

To support this Challenge, we introduce the Children Gait Pose Sequence (CGPS) dataset, which features thousands of 60 FPS 2D human keypoint sequences capturing the natural walking behaviors of children across various age groups (2 to 16 years old), as detailed in~\Cref{sec:dataset}. With the first available dataset for such tasks and a comprehensive ground truth for each track, the Challenge enables participating teams to effectively evaluate their solutions. 

The Challenge will be launched with two specific problem tracks: \textbf{Edinburgh Visual Gait Score (EVGS) Scoring} and \textbf{classification of Gait Patterns in Bilateral Spastic Cerebral Palsy}. Each track is chosen in consultation with expert pediatricians based on the clinical value of potential solutions, and they are all interconnected in some way~\cite{yun2023selective}. \looseness=-1

This paper presents a comprehensive summary of the preparation and results of the 1st AI Children Challenge. The following sections describe the challenge setup~\Cref{sec:challenge}, dataset preparation~\Cref{sec:dataset}, evaluation methodology~\Cref{sec:evaluation}, results and summary of participant submissions~\Cref{sec:results}, and conclude with a discussion of the findings and future research directions~\Cref{sec:conclusion}.
\section{Challenge Setup}
\label{sec:challenge}

\begin{table}[!t]
    \centering
    \caption{\textbf{17 Scoring Items of EVGS.} The Edinburgh Visual Gait Score (EVGS) is a clinical tool developed to visually assess gait deviations in ambulatory children using coronal and sagittal video recordings. It evaluates 17 observational scoring items for each limb that are graded on a three-point ordinal scale.}
    \vspace{-2mm}
    \label{tab:items}
    \resizebox{\linewidth}{!}{ 
    \begin{tabular}{clcccc}
        \toprule
        \textbf{No.} & \textbf{Gait Pattern} & \textbf{Abbr.} & \textbf{Score Scale} & \textbf{View}\\
        \midrule
        1 & Initial Contact in Stance & IC & 0, 1 & Sagittal \\
        2 & Heel Lift in Stance & HL & 0, 1 & Sagittal \\
        3 & Max Ankle Dorsiflexion in Stance & SAD & 0, 1 & Sagittal \\
        4 & Hind-foot Varus/Valgus in Stance & HVV & 0, 1 & Coronal \\
        5 & Foot Rotation in Stance & FRT & 0, 1 & Coronal \\
        6 & Foot Clearance in Swing & FCL & 0, 1 & Sagittal \\
        7 & Max Ankle Dorsiflexion in Swing & WAD & 0, 1 & Sagittal \\
        8 & Knee Progression Angle in Mid-Stance & KPA & 0, 1 & Coronal \\
        9 & Peak Knee Extension in Stance & KEX & 0, 1 & Sagittal \\
        10 & Knee Position in Terminal Swing & KPS & 0, 1 & Sagittal \\
        11 & Peak Knee Flexion in Swing & KFX & 0, 1 & Sagittal \\
        12 & Peak Hip Extension in Stance & HEX & 0, 1 & Sagittal \\
        13 & Peak Hip Flexion during Swing & HFX & 0, 1 & Sagittal \\
        14 & Pelvic Obliquity at Mid-Stance & POB & 0, 1 & Coronal \\
        15 & Pelvic Rotation at Mid-Stance & PRT & 0, 1 & Coronal \\
        16 & Peak Sagittal Trunk Position in Stance & TSG & 0, 1 & Sagittal \\
        17 & Maximum Trunk Lateral Shift & TLT & 0, 1 & Coronal \\
        \bottomrule
    \end{tabular}
    }
\end{table}

\begin{figure}[!t]
    \centering
    \includegraphics[width=0.99\linewidth]{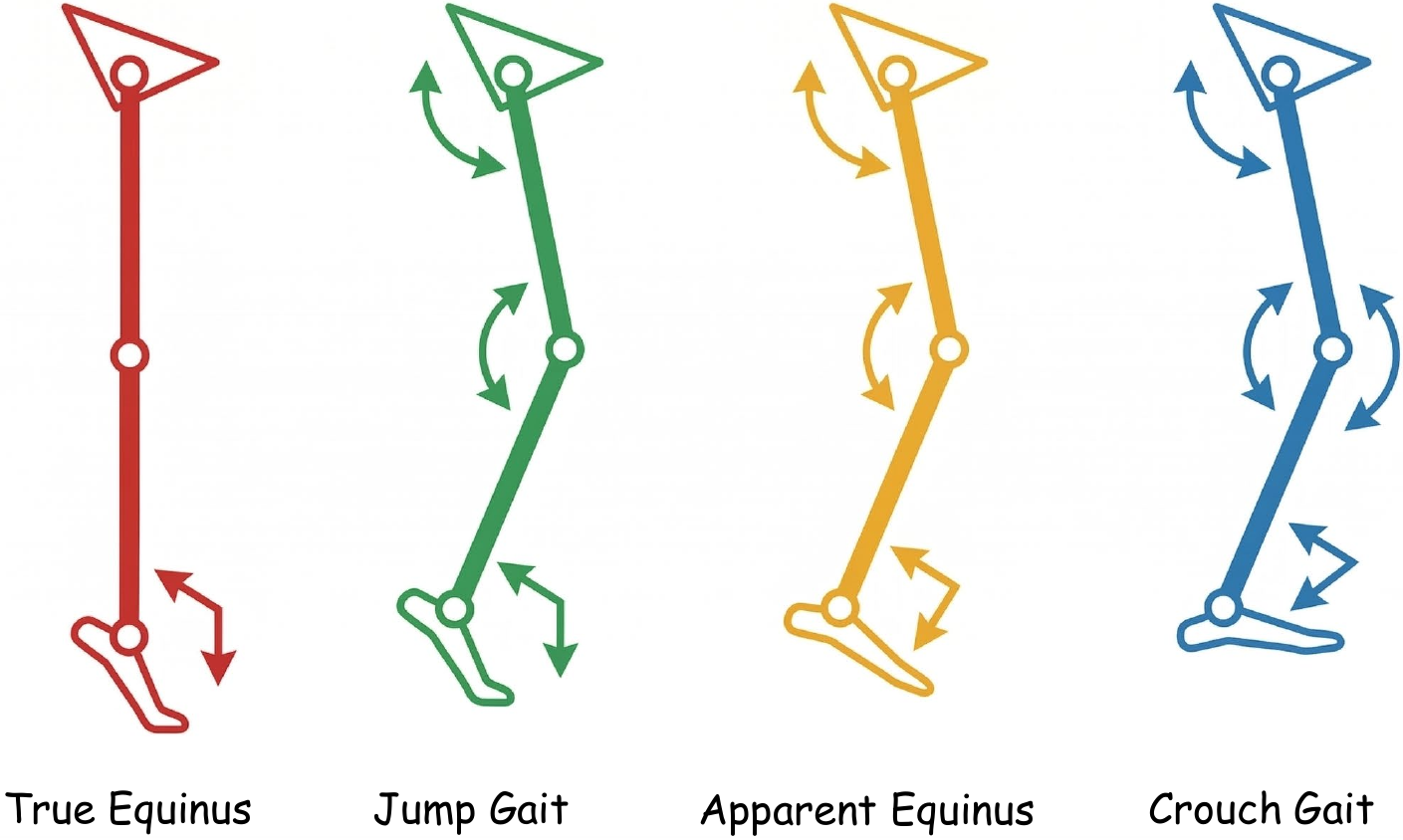}
    \vspace{-2mm}
    \caption{\textbf{The Summary of the Classification of Gait Patterns in Bilateral Spastic Cerebral Palsy.} From left to right are four different gait patterns in bilateral spastic CP: \textbf{True Equinus (red):} Defined by hip and knee extension (sometimes knee recurvatum) and an equinus foot position. \textbf{Jump Gait (green):} Characterized by hip and knee flexion with an equinus foot position, typically associated with anterior pelvic tilt and lumbar lordosis. \textbf{Apparent Equinus (yellow):} Features increased hip and knee flexion and a decreased equinus ankle angle, leading to increased dorsiflexion. \textbf{Crouch Gait (blue):} Marked by excessive hip and knee flexion, scissoring, and excessive dorsiflexion. \looseness=-1}
    \label{fig:bilateral}
    \vspace{-5mm}
\end{figure}

The 1st AI Children Challenge will be held on Kaggle~\cite{li2026kaggle}. It allows participants to compete in one or all of the following two tracks. The Challenge follows a structured timeline, with the Challenge launching and registration opening on March 23, 2026. Participants are required to submit their challenge results by April 23, 2026. For teams intending to publish accompanying papers, the workshop paper (non-proceeding track) submission deadline was set for May 15, 2026. All deadlines are at 11:59 PM (Anywhere on Earth) of the corresponding day unless otherwise stated.

To promote transparency and reproducibility, teams competing for top rankings were required to publicly release their code for the launched post-evaluation. This policy ensured that leaderboard results could be independently verified and contributed to the broader research community.

\begin{figure*}[!t]
    \centering
    \includegraphics[width=0.99\linewidth]{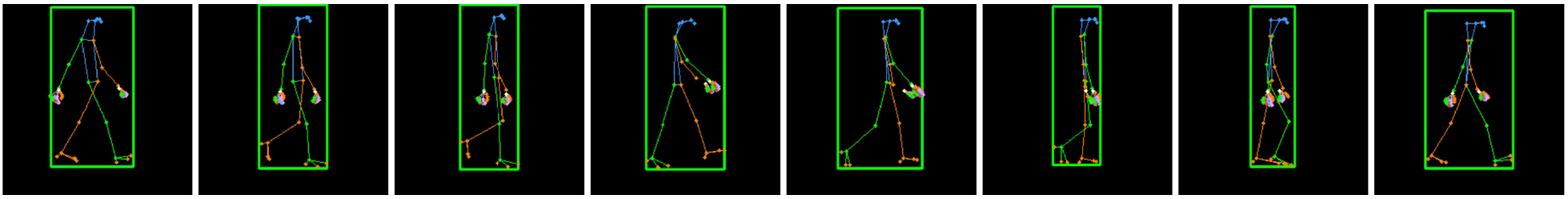}
    \vspace{-2mm}
    \caption{\textbf{Dataset Examples Visualization.} The CGPS dataset provides keypoint sequences and bounding boxes per frame. We use SAM3~\cite{carion2025sam} to perform instance segmentation and Sapiens-2B~\cite{khirodkar2024sapiens} to perform pose estimation to obtain bounding boxes and keypoints. We will release the keypoint sequences for the participants.}
    \label{fig:visualization}
\end{figure*}

\paragraph{Track 1 - EVGS Scoring.} As shown in~\Cref{tab:items}, Edinburgh Visual Gait Score (EVGS)~\cite{read2003edinburgh} is a kind of observational scale which is critical for diagnosing neurodevelopmental disorders such as cerebral palsy (CP). Therefore, designing dedicated pediatric modeling techniques and developing a model capable of automatic scoring to provide pediatricians with a reference is essential for clinical diagnosis. In this track, participants are tasked with developing algorithms capable of conducting fine-grained kinematic analysis to accurately predict EVGS metrics directly from provided children's 2D keypoint sequences. Evaluation of Track 1 is based on distinguishing subtle kinematic differences, measured by Accuracy and RMSE, which tells the accuracy of the predictions and the difference between the predicted score and the ground truth for each item.

\paragraph{Track 2 - Classification of Gait Patterns in Bilateral Spastic Cerebral Palsy.} Although Gait classification systems for children with bilateral lower limb spasticity have been developed, there are limitations to their clinical relevance and applicability in orthotic prescription. Studies have shown low levels of validity and reliability for these classification systems~\cite{dobson2007gait}, emphasizing the need for comprehensive approaches that address the range and magnitude of gait deviations in children with bilateral spastic CP. A summary of the classification of gait patterns in bilateral spastic CP is shown in~\Cref{fig:bilateral}. As a track running parallel to Track 1, this track focuses on dealing with this issue, which challenges teams to classify specific gait patterns of bilateral spastic CP by recognizing subtle, pathological variations in children's gait behaviors. Evaluation of Track 2 is based on the clinical diagnostic performance, measured by the accuracy rate and the F1 score.
\section{Children Gait Pose Sequence Dataset}
\label{sec:dataset}

\begin{figure*}[!t]
    \centering
    \begin{minipage}[c]{0.25\textwidth}
        \centering
        \begin{minipage}[c]{0.82\textwidth}
            \centering
            \includegraphics[width=\textwidth]{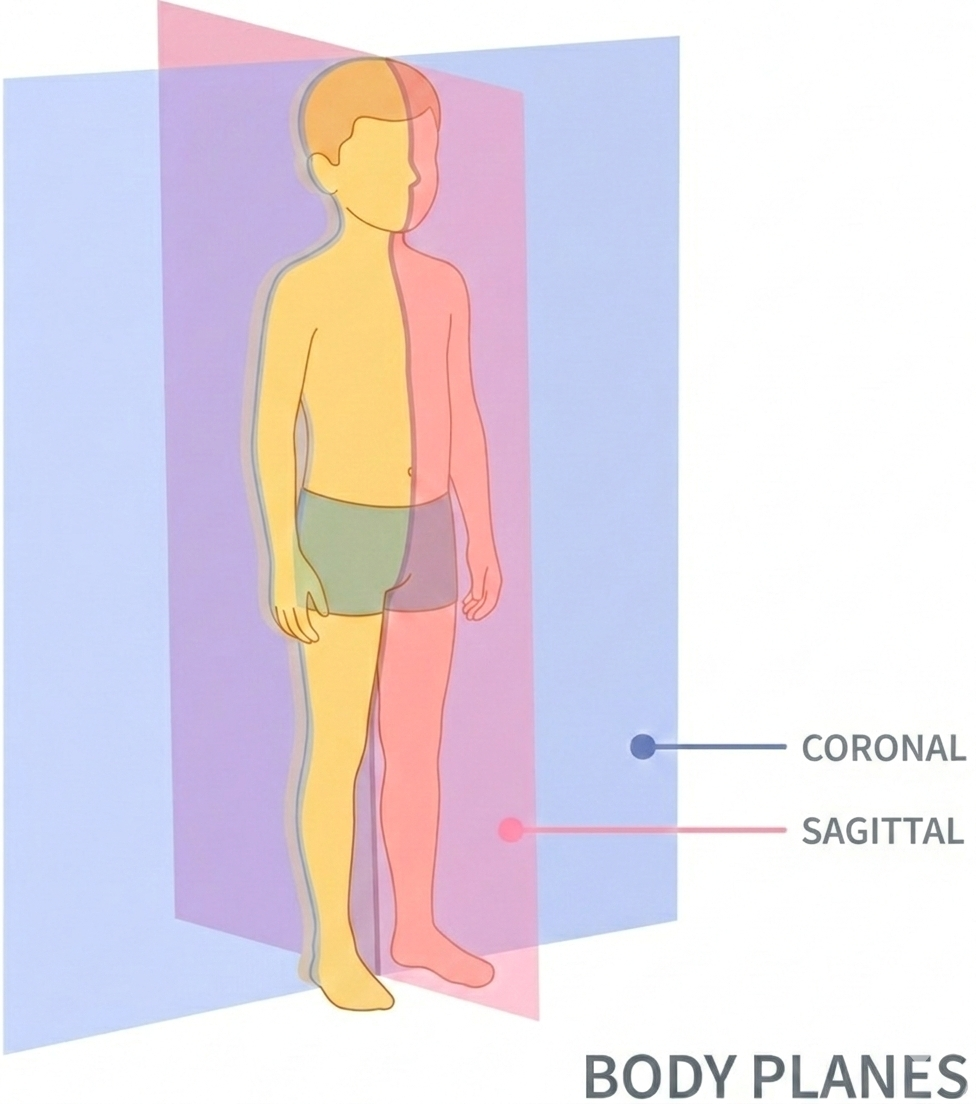}
        \end{minipage}
        
    \end{minipage}
    \hfill
    \begin{minipage}[c]{0.74\textwidth}
        \centering
        \begin{minipage}[c]{\textwidth}
            \centering
            \includegraphics[width=\textwidth]{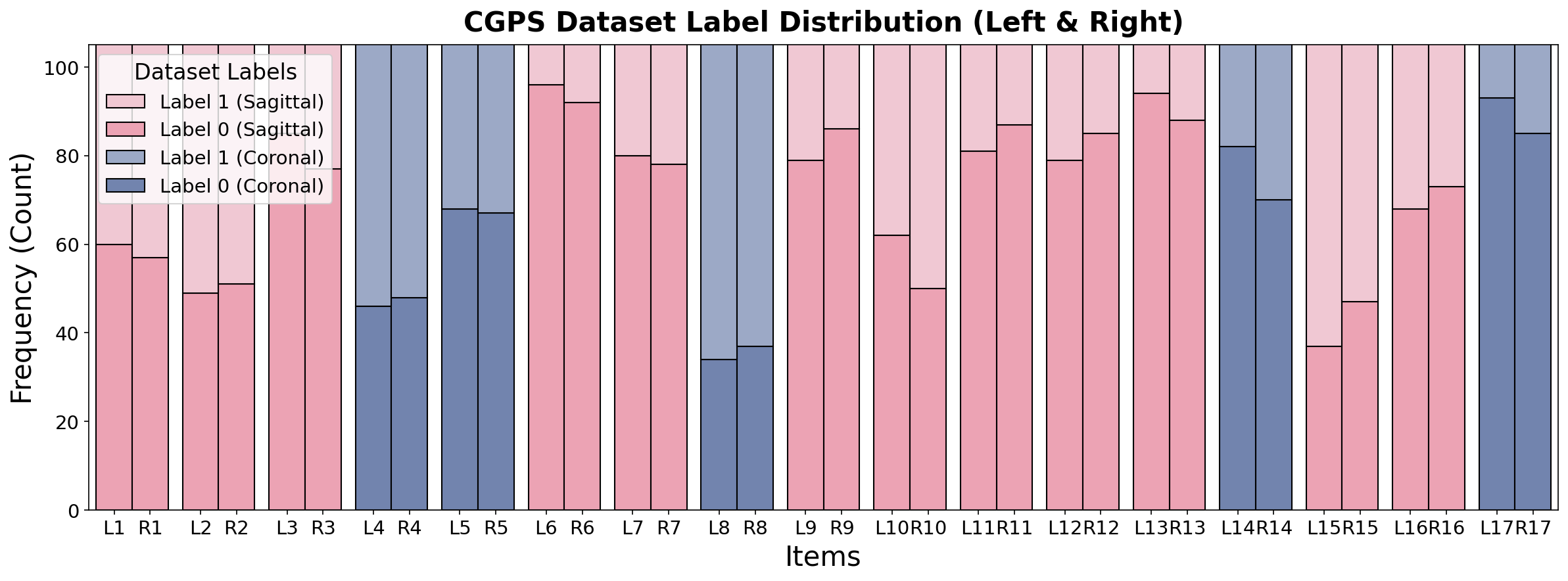}
        \end{minipage}
        
    \end{minipage}
    \vfill
    \begin{minipage}[c]{\textwidth}
        \centering
        \includegraphics[width=\textwidth]{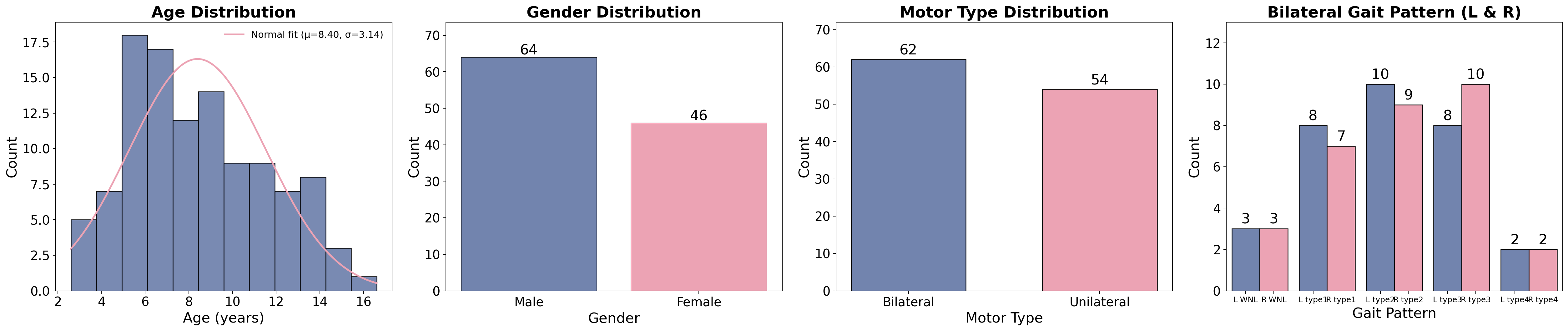}
    \end{minipage}
    \vspace{-2mm}
    \caption{\textbf{The Overview of our CGPS Dataset.} The CGPS dataset comprises 110 pediatric patients, with each evaluation covering a total of $2\times17$ EVGS scoring items (17 items per limb) across two body planes, a CP motor type, and a corresponding gait pattern. \textit{Top Left:} The child's body planes. \textit{Top Right:} The label distribution of the CGPS dataset. \textit{Bottom:} From left to right: age distribution, gender distribution, CP motor type distribution, and bilateral spastic CP gait patterns distribution. Overall, the dataset maintains a nearly balanced profile, providing the necessary diversity for training or evaluation in visual gait analysis.}
    \label{fig:CGPS_overview}
    \vspace{-5mm}
\end{figure*}

We introduce the Children Gait Pose Sequence (CGPS) dataset to support this Challenge. Before the data annotation and model design, the IRB approval is obtained from affiliated hospitals. The dataset contains 2D keypoints annotation per frame across a total number of 339,236 frames (1,185 videos) of size $1920\times 1080$ (2K) with 17 EVGS~\cite{read2003edinburgh} sub-item annotations per limb and the patient-level diagnosis results of cerebral palsy (CP) motor types (Unilateral or Bilateral) and corresponding gait patterns. The videos were recorded at Shenzhen Children's Hospital using smartphone cameras and action cameras, positioned simultaneously to capture sagittal and coronal views. We then used SAM3~\cite{carion2025sam} to detect bounding boxes and Sapiens-2B~\cite{khirodkar2024sapiens} to detect 2D keypoint sequences, where all annotations are manually adjusted by human annotators. \Cref{fig:visualization} shows examples of videos and annotations. In total, 110 patients participated in the data collection. In some cases, multiple recordings per view are available. The dataset provides detailed annotations per frame and per video, including the subject’s body bounding box (detected with SAM 3~\cite{carion2025sam} and manually selected by human annotators), the 2D human keypoint sequences (detected with Sapiens-2B~\cite{khirodkar2024sapiens} and manually adjusted by human annotators), the EVGS sub-items, CP motor types and corresponding gait patterns (annotated by an experienced pediatricians in the author team and reviewed by a senior pediatrician author with 40 years of clinical experience). \Cref{fig:CGPS_overview} shows the EVGS scores distribution, CP motor types distribution, and bilateral spastic CP gait patterns distribution among all patients.

To capture the dataset, a primary camera was positioned at the terminus of an 8-meter walkway to record the coronal (frontal and posterior) view. A secondary camera was oriented orthogonally, facing the center of the walkway, to capture the sagittal (lateral) view. This lateral camera was positioned at a sufficient distance to ensure its field of view encompasses the middle four meters of the trial space. This specific distance was calibrated to guarantee the capture of 2-3 complete gait cycles (strides) per subject. Then the videos were further processed to obtain bounding boxes and keypoint sequences per frame. \Cref{tab:items} details the 17 fine-grained gait parameters annotated in the CGPS dataset, which strictly adhere to the EVGS reference guide to ensure high clinical validity. \Cref{fig:bilateral} details the four gait patterns in bilateral spastic CP. While the standard EVGS protocol employs a three-point ordinal scale (0: normal, 1: moderate deviation, 2: severe deviation) for each parameter, the natural distribution of pediatric gait pathologies inherently results in a severe class imbalance, particularly for the most extreme deviations. To mitigate this imbalance and establish a robust computational benchmark, we binarized the assessment by merging scores 1 and 2. Consequently, the prediction for each fine-grained gait parameter is formulated as a binary classification task (see \Cref{fig:CGPS_overview}) discriminating between ``typical'' and ``atypical'' gait patterns. \looseness=-1

The age and gender distribution of the patients are shown in~\Cref{fig:CGPS_overview}. The age histogram is overlaid with a fitted normal distribution curve to illustrate the central tendency of the patient demographics. The participants span a critical developmental window from 2.6 to 16.6 years old, with a mean age of $\mu=8.40, \sigma=3.14$ years, which is highly clinically relevant and highlights a primary focus on the pediatric population. It also details the gender composition of the dataset. The patients consist of 58.2\% male and 41.8\% female subjects. This nearly balanced gender distribution helps to prevent the model from learning shortcuts for specific genders, ensuring sufficient diversity and fairness for training and evaluating models for visual gait analysis.

To prevent identity leakage and shortcut bias, we implement a strict object-level data split based on the unique \textbf{Patient ID} in the experiment. For different tracks, the test set is randomly selected, and we ensure all annotations (scores distribution in Track 1 and gait patterns distribution in Track 2) are sampled nearly balanced. All sequences, corresponding annotations, and derived gait cycles for any given patient belong exclusively to either the training or the test set. For different tracks, we select different sets of patients as test sets and will only release sequences corresponding to the track and specific annotations.

\section{Evaluation Protocols}
\label{sec:evaluation}

The 1st AI Children Challenge features two tracks spanning EVGS scoring and gait patterns classification. The datasets, task objectives, submission formats, and evaluation metrics are summarized below.

\begin{figure*}[!t]
    \centering
    \begin{lstlisting}[caption={\textbf{Required Submission Format.} Participants should fill in the cells under the requirements, and leave the blank cells with \textbf{-1}.}, label={lst:csv}]
    ID,L1,L2,L3,L4,L5,L6,L7,L8,L9,L10,L11,L12,L13,L14,L15,L16,L17,R1,R2,R3,R4,R5,R6,R7,R8,R9,R10,R11,R12,R13,R14,R15,R16,R17,Total,Left_gait_subtype,Right_gait_subtype
    track1-4,...
    ...
    track2-4,...
    ...
    \end{lstlisting}
    \vspace{-5mm}
\end{figure*}

\subsection{Track 1 Evaluation}

Track 1 challenges teams to conduct fine-grained kinematic analysis and accurately predict 34 EVGS item scores from the provided multi-view annotations. The core of this evaluation focuses on distinguishing subtle kinematic differences, which are measured by the discrepancy between the predicted score and the clinical ground truth for each EVGS item. The dataset covers the entire CGPS dataset, carefully selecting the evaluation set to ensure that the number of positive and negative samples is nearly equal, ensuring the accuracy of the test. The ratio of the training set to the evaluation set is 94:16. \looseness=-1

Each patient provides multi-view JSON files, including 2D keypoint sequences and bounding boxes per frame, and the EVGS scores per patient. Participants are required to design algorithms and models to track the correct patients, analyze the gait features, and score the EVGS items based on the criteria automatically.  

In this track, we use Accuracy and Root Mean Square Error (RMSE) to comprehensively evaluate the predictive ability of the fine-grained kinematic analysis and EVGS scoring of the model. Specifically, for each patient in the evaluation set, we will calculate the overall accuracy:

\begin{equation}
    \mathrm{Accuracy}_{\mathrm{track1}} = \frac{1}{N}\sum^{N}_{i=1}\mathbf{1}\left[y_i= \hat{y}_i\right],
\end{equation}
where $N$ is the total number of the EVGS scoring items across all patients in the evaluation set of Track 1, $\hat{y}_i$ is the predicted score of a specific EVGS item, and $y_i$ is the corresponding clinical ground truth, $\hat{y}_i, y_i\in \{0, 1\}$. 

Besides, we calculate the patient-level error. The error between the ground truth and the predicted values is calculated using the Root Mean Square Error (RMSE):

\begin{equation}
    \mathrm{RMSE} = \sqrt{\frac{1}{M}\sum_{i=1}^{M}(p_{i} - g_{i})^{2}},
\end{equation}
where $M$ is the total number of sampled patients in the evaluation set of Track 1, $p_i = \sum\limits_j \hat{y}_{i,j}$ is the sum of predicted EVGS item scores for a particular patient $i$, and $g_i = \sum\limits_j y_{i,j}$ is the sum of corresponding clinical ground truth. 

To ensure that a smaller gap translates to a higher score, the score for Track 1, denoted as $S_{1}$, uses the Accuracy and normalized RMSE (NRMSE) with equal weights to scale the results relative to all participating teams: \looseness=-1
 
\begin{equation}
    \mathrm{NRMSE} = \frac{\mathrm{RMSE}}{34},
\end{equation}
\begin{equation}
    S_1 = \frac{\mathrm{Accuracy}_\mathrm{Track1} + 1 - \mathrm{NRMSE}}{2}.
\end{equation}

This scoring mechanism effectively rewards models that closely align with the precise ground-truth measurements. 

\subsection{Track 2 Evaluation}

Track 2 evaluates the model's clinical diagnostic performance in accurately classifying specific gait patterns of bilateral spastic CP. This task challenges algorithms to recognize subtle, pathological variations in children's gait behaviors that dictate different bilateral lower limb spasticity classifications. We select a subset from the CGPS dataset consisting exclusively of patients with bilateral spastic CP. The evaluation set is also carefully selected to ensure that the number of five gait pattern samples is nearly equal. The ratio of the training set to the evaluation set is 22:9.

Similar to Track 1, each patient provides multi-view JSON files, including 2D keypoint sequences and bounding boxes per frame, and detailed gait patterns per patient. Participants are required to design algorithms and models to track the correct patients, analyze the gait features, and score the EVGS items based on the criteria automatically. 

To rank participant submissions, we use Accuracy and F1 Score, designed to balance recall and precision. This ensures that models do not over-predict specific types (leading to low precision) nor under-predict specific types (resulting in low recall). For this challenge, we use the macro-averaged F1 score for each patient, which is particularly suited for multi-label classification problems. This approach ensures a fair evaluation of each image independently, rather than being influenced by class distribution. For each gait pattern, the F1 score is computed as

\begin{equation}
    \mathrm{F1}_k = \frac{2P_kR_k}{P_k + R_k},
\end{equation}
where for each gait pattern in bilateral spastic CP $k\in\mathcal{B}, \mathcal{B}=\{\mathrm{TE, JG, AE, CG, WNL}\}$, we consider it as the positive class, and the other four categories as the negative classes. Therefore, the macro-averaged F1 score is:

\begin{equation}
    \mathrm{F1}_{\mathrm{macro}} = \frac{1}{|\mathcal{B}|}\sum_{k\in\mathcal{B}}\mathrm{F1}_k.
\end{equation}

We also calculate the overall accuracy via

\begin{equation}
    \mathrm{Accuracy}_\mathrm{Track2} = \frac{1}{2M}\sum_{k\in\mathcal{B}}\mathrm{TP}_k,
\end{equation}
where $M$ is the number of sampled patients in the evaluation set of Track 2 (yielding $2M$ evaluation targets). Notice that the evaluation set is different from that of Track 1.

Denote the evaluation score of Track 2 as $S_2$, which is calculated by aggregating the macro-averaged F1 score and Accuracy with equal weights, reflecting holistic clinical utility and robustness:

\begin{equation}
    S_2 = \frac{\mathrm{Accuracy}_{\mathrm{Track2}} + \mathrm{F1}_\mathrm{macro}}{2}.
\end{equation}

\subsection{Evaluation System}

The 1st AI Children Challenge uses a centralized evaluation portal via Kaggle~\cite{li2026kaggle}. Participants are asked to create an evaluation account with an email and password and verify their email to activate the account. Once registered, teams are then allowed to submit their predictions. 

We just set up one overall leaderboard that combines the scores of two tracks. During the Challenge, teams could see a leaderboard with the best results from each team. The rankings are based on the total scores from both tracks, and the scores of the two tracks each account for 50\% to determine the final score $S_\mathrm{Kaggle}$, which is calculated as

\begin{equation}
 S_\mathrm{Kaggle} = \frac{S_1 + S_2}{2}.
\end{equation}

The submission file should be in CSV format, which contains both track results. The file should contain a header. \Cref{lst:csv} shows the required format for submission. Participants should use \texttt{track1-} and \texttt{track2-} as prefixes, and then add the patient ID to distinguish between Track 1 and 2. For cells that do not belong to the current track or the track that they do not participate in, participants should enter \textbf{-1}. The results of Track 1 should be listed first, followed by Track 2, and the patient IDs should be sorted in ascending order in the first column. For Track 1, the cell should be filled with \textbf{0 or 1}, indicating ``typical'' or ``atypical''. For Track 2, the cell should be filled with one in \textbf{type1, type2, type3, type4, WNL}, which represents the four gait patterns in bilateral spastic CP and \textit{Within Normal Limits}.

To qualify for awards via the Kaggle leaderboard, teams are required to: \looseness=-1

\begin{itemize}
    \item Cannot enter or submit from multiple accounts.
    \item Publicly release reproducible code, train models before the deadline to verify the submission, promote transparency and reproducibility.
    \item Write and submit a technical report to the CV4CHL Workshop non-proceeding track.
\end{itemize}

Submission limits were set as follows:

\begin{itemize}
    \item Up to 10 submissions per day (Anywhere on Earth).
    \item Only select 1 final submission for judging.
\end{itemize}

After the Challenge on Kaggle closes, we will launch post-evaluation and re-evaluate the submissions locally using the released code and model from teams. Any team that has not made its code public, or whose results differ significantly from the local evaluation, will be deemed non-compliant and forfeit its eligibility for awards. The rankings of the post-evaluation are determined by averaging the three evaluation scores:

\begin{equation}
    S_\mathrm{post} = \frac{1}{3}\sum_{i=1}^{3} 0.7S_{i, 1} + 0.3S_{i, 2},
\end{equation}
where $S_{i,\{1,2\}}$ denotes the i-th evaluation for Track 1 or 2.

The three evaluations are as follows: a test using the same training and test sets as those published on Kaggle~\cite{li2026kaggle}, and two additional evaluations using differently partitioned training and test sets. All training and evaluation are performed on a single NVIDIA L40S GPU, and we ensure that all test hardware configurations are the same.

\section{Challenge Results}
\label{sec:results}

\begin{table}[!t]
    \centering
    \caption{\textbf{Leaderboard on Kaggle.} We report the top 10 teams' results of their final score $S_\mathrm{Kaggle}$ and entries as of the Challenge closes. The complete ranking list can be viewed at~\cite{li2026kaggle}.}
    \vspace{-2mm}
    \label{tab:rank}
    \resizebox{0.7\linewidth}{!}{
    \begin{tabular}{cccc}
        \toprule
        \textbf{Rank} & \textbf{Team} & \textbf{Score} & \textbf{Entries} \\
        \midrule
        1 & seantangth & 0.92989 & 238 \\
        2 & Sardor Razikov & 0.90271 & 40 \\
        3 & Kun Zhang & 0.89510 & 105 \\
        4 & yingfali & 0.87747 & 58 \\
        5 & Yutong He 2025 & 0.87486 & 96 \\
        6 & Zihan Ji & 0.85954 & 63 \\
        7 & Master Teo & 0.83171 & 46 \\
        8 & HUN UUNNN & 0.82690 & 60 \\
        9 & CCE & 0.76738 & 73 \\
        10 & PaulThangNguyen & 0.73642 & 3 \\
        \bottomrule
    \end{tabular}}
\end{table}

Overall, the 1st AI Children Challenge attracted 57 teams from around the world, with over 100 participants. 7 teams participated in our post-evaluation and submitted technical reports introducing their solutions. \Cref{tab:rank} provides a summary of the top 10 teams' results on Kaggle as of the Challenge closes. \Cref{tab:post} shows the post-evaluation results of every single evaluation and the overall final scores. \Cref{tab:track1} and~\Cref{tab:track2} provide the post-evaluation results for Tracks 1 and 2. Detailed summary of Track 1 and 2 can be found in~\Cref{sec:track1_sum} and~\Cref{sec:track2_sum}.

\begin{table}[!t]
    \centering
    \caption{\textbf{Post-Evaluation Results.} We report the post-evaluation results of each evaluation and the final score.}
    \vspace{-2mm}
    \label{tab:post}
    \resizebox{0.99\linewidth}{!}{ 
    \begin{tabular}{cccccc}
        \toprule
        \textbf{Rank} & \textbf{Team} & \textbf{Eval 1} & \textbf{Eval 2} & \textbf{Eval 3} & \textbf{Average} \\ 
        \midrule
        1 & seantangth & 0.836046 & 0.586667 & 0.658576 & 0.693763 \\
        2 & CCE & 0.749006 & 0.632553 & 0.614234 & 0.665264 \\
        3 & Zihan Ji & 0.635633 & 0.642793 & 0.610913 & 0.629780 \\
        4 & Btaji Crew & 0.613517 & 0.536837 & 0.642021 & 0.597458 \\
        5 & Yutong He 2025 & 0.507284 & 0.634737 & 0.648482 & 0.596834 \\
        6 & Sardor Razikov & 0.507429 & 0.634125 & 0.610588 & 0.584047 \\
        7 & Anqi Wang & 0.422472 & 0.494943 & 0.546764 & 0.488060 \\
        \bottomrule
    \end{tabular}}
\end{table}

\subsection{Summary for Track 1}
\label{sec:track1_sum}

Track 1 focused on automated EVGS scoring for pediatric patients. Participants were tasked with developing algorithms to conduct fine-grained kinematic analysis and accurately predict EVGS directly from multi-view 2D children's keypoint sequences. Emphasis was placed on distinguishing subtle kinematic differences inherent in the developing motor system and overcoming the limitations of adult-centric foundation models. The top submissions reflected dominant strategies in handling high intra-class variance and extracting clinically guided features from data.

The winning team, CCE, proposed a clinically grounded multi-view pipeline combining pretrained deep motion representations with hand-crafted kinematic features. Their method used MotionBERT~\cite{zhu2023motionbert} with PCA compression, EVGS-aligned clinical descriptors, and a multi-output ensemble, enabling accurate prediction of the 34-item scores.

Team Btaji Crew achieved EVGS predictions by extracting explicit biomechanical descriptors, such as joint angles and posture, from pose sequences. These clinical features were then classified using an optimized ensemble of Gradient Boosting, Extra Trees, and Random Forest models.

Team seantangth proposed a hybrid approach that combined machine learning with deterministic clinical consistency rules. To predict individual EVGS items, their method leveraged 2D pose-derived gait features alongside selected ML models and rule-based constraints. This integration ensured that the final deterministically assembled predictions maintained strict clinical validity.

Team Zihan Ji developed PoseActionTransformer, which first extracted inter-keypoint postural features per frame via a spatial Transformer, and then aggregated gait dynamics through a temporal Transformer. To generate the final results, the system combined the scores from the four camera views via weighted soft voting, assigning weights based on the clinical geometric relevance of each view to the specific EVGS item. \looseness=-1

Team Sardor Razikov combined consensus voting across 13 independent pose estimation models with type-conditional statistical priors. The base predictions were refined using clinical type-informed corrections, adjusting scores that deviated from expected EVGS prototypes, while incorporating specialized handling for asymmetric limb presentations.

Team Yutong He 2025 utilized a Dual-stream Spatio-temporal Transformer (DSTformer) to process stabilized 2D keypoint sequences. The architecture extracted kinematic features using parallel spatial and temporal attention streams, dynamically merged by an Adaptive Attentional Fusion module, and generated final EVGS predictions through specialized regression heads combined with a multi-view voting mechanism.

\begin{table}[!t]
    \centering
    \caption{\textbf{Track 1 Results.} }
    \vspace{-2mm}
    \label{tab:track1}
    \resizebox{0.99\linewidth}{!}{ 
    \begin{tabular}{cccccc}
        \toprule
        \textbf{Rank} & \textbf{Team} & \textbf{Eval 1} & \textbf{Eval 2} & \textbf{Eval 3} & \textbf{Average} \\ 
        \midrule
        1 & CCE & 0.76673 & 0.75936 & 0.80843 & 0.77817 \\
        2 & Btaji Crew & 0.75953 & 0.76691 & 0.80244 & 0.77629 \\
        3 & seantangth & 0.76578 & 0.76418 & 0.76738 & 0.76578 \\
        4 & Zihan Ji & 0.71909 & 0.78256 & 0.73139 & 0.74435 \\ 
        5 & Sardor Razikov & 0.62823 & 0.77079 & 0.7756 & 0.72487 \\
        6 & Yutong He 2025 & 0.65993 & 0.71067 & 0.73571 & 0.70210 \\
        7 & Anqi Wang & 0.48639 & 0.58545 & 0.71633 & 0.59606 \\
        \bottomrule
    \end{tabular}}
\end{table}

\subsection{Summary for Track 2}
\label{sec:track2_sum}

Track 2 evaluated the clinical diagnostic performance of models in accurately classifying specific gait patterns of bilateral spastic cerebral palsy. This task challenged algorithms to recognize subtle, pathological variations in children's gait behaviors that dictate different bilateral lower limb spasticity classifications. Participants were required to analyze the gait features directly from the provided multi-view 2D keypoint sequences. The top submissions reflected dominant strategies in handling severe class imbalance and leveraging clinical constraints or cross-track signals to overcome the limited sample size.

The winning team, seantangth, utilized an XGBoost subtype classifier for their predictions. Their approach was specifically refined by integrating Rodda \& Graham's clinical constraints alongside left/right symmetry rules to ensure the final classification maintained strict clinical consistency.

Team CCE used a side-specific limb representation to increase the effective number of training samples. Their approach combined sample-efficient machine learning classifiers with a compact Conv-BiGRU model. Furthermore, they incorporated cross-track clinical knowledge by deriving a Rodda-rule prior from their Track 1 EVGS predictions, which was seamlessly integrated with the Track 2 model outputs through a light-gating strategy.

Team Zihan Ji tackled the subtype classification task by transferring their Track 1 backbone to a five-class classifier. They fine-tuned this model using a multiclass focal loss design, class-prior initialization, and stronger pose augmentations to handle the small labeled cohort.

Team Yutong He 2025 extended their stabilized Dual-stream Spatio-temporal Transformer (DSTformer) framework with a dedicated one-shot gait classification head to specifically handle class imbalance. This was culminated with a multi-view voting mechanism to derive reliable final subtype predictions. \looseness=-1

Team Sardor Razikov exploited the patient overlap between the training sets as a cross-track structural signal. They built per-type EVGS prototype vectors from the training overlap and assigned each test patient's subtype using an L2 nearest prototype matching strategy, effectively bypassing complex temporal modeling for the small-sample classification task. \looseness=-1

Team Btaji Crew approached the problem using a compact temporal convolutional network. Operating directly on multi-view 2D pose sequences, their model processed fixed-length pose tensors and utilized separate left and right classification heads to capture subtype dynamics efficiently.

\begin{table}[!t]
    \centering
    \caption{\textbf{Track 2 Results.} }
    \vspace{-2mm}
    \label{tab:track2}
    \resizebox{0.99\linewidth}{!}{ 
    \begin{tabular}{cccccc}
        \toprule
        \textbf{Rank} & \textbf{Team} & \textbf{Eval 1} & \textbf{Eval 2} & \textbf{Eval 3} & \textbf{Average} \\ 
        \midrule
        1 & seantangth & 1 & 0.17247 & 0.4047 & 0.52572 \\
        2 & CCE & 0.70765 & 0.33667 & 0.16111 & 0.40181 \\
        3 & Zihan Ji & 0.4409 & 0.31667 & 0.3298 & 0.36246 \\
        4 & Yutong He 2025 & 0.15111 & 0.45756 & 0.44495 & 0.35121 \\ 
        5 & Sardor Razikov & 0.22556 & 0.31524 & 0.22556 & 0.25545 \\
        6 & Anqi Wang & 0.27333 & 0.28376 & 0.15111 & 0.23607 \\
        7 & Btaji Crew & 0.27282 & 0 & 0.26771 & 0.18018 \\
        \bottomrule
    \end{tabular}}
\end{table}
\section{Discussion and Conclusion}
\label{sec:conclusion}

The 1st AI Children Challenge marks a significant milestone in advancing applied computer vision across child healthcare, child education, and pediatrics. With newly proposed datasets and novel benchmarks, the Challenge pushed the boundaries of fine-grained children's gait analysis.

Track 1 introduces clinical annotations of professional pediatric doctors' EVGS scores for the CGPS dataset. This is an asset of critical importance for pediatric diagnostics. A key research opportunity remains in automated severity assessment, especially in handling subtle pathological cues and high inter-patient variability in pediatric cases. Future iterations may further challenge generalization with increased clinical complexity, such as varied imaging conditions or diverse anatomical appearances.

Track 2 shifts the analytical focus toward bilateral spastic, a specific motor type of cerebral palsy, by systematically investigating its four distinct sub-gait patterns. A key research opportunity remains in the robust differentiation of transitional or borderline gait patterns, especially when handling noisy data and atypical clinical presentations. Future iterations may further challenge generalization by expanding to other CP motor types or incorporating longitudinal data to evaluate gait evolution over time.

The 1st AI Children Challenge highlights a strong trend toward clinically-guided feature extraction, fine-grained kinematic modeling, and diagnostic interpretability. As AI systems move toward deployment in real-world clinical workflows, the continued convergence of computer vision, biomechanics, and medical expertise will drive innovation in precision pediatric healthcare.

\section{Acknowledgement}
\label{sec:acknowledgement}

The CGPS dataset is developed through extensive data curation efforts enabled by close collaboration. Key contributors include Shenzhen Children's Hospital and PediaMed AI, alongside academic partners such as the University of Illinois Urbana-Champaign and The Hong Kong Polytechnic University. The Challenge also benefited from thorough review efforts by researchers. We would like to thank the gait research community for their previous work. We thank the broader research community for their participation, which helps establish the AI Children Challenge as a leading benchmark in pediatric healthcare. The contributions of participants, reviewers, and collaborators are essential to the success of the Challenge.

Furthermore, we are deeply grateful to the participants of the 1st AI Children Challenge for their innovative solutions. We especially acknowledge the top-performing teams and their members for contributing their methodological details to this extended report: Jen-Wei Kuo and Yi-Ting Ku (National Cheng Kung University), Muhammad Ibrahim Qasmi (Government Graduate College Sahiwal), Tze-Hsiang Tang (Schneider Electric Taiwan), Zihan Ji and Jingkai Bao (University of Illinois Urbana-Champaign), Hongtao Sun and Chensong Wang (Nanjing University), Yisu Li (The Chinese University of Hong Kong), Sardor Razikov (Independent Researcher), and Yutong He (Beijing University of Posts and Telecommunications). Their dedication and open-source contributions are vital to establishing this challenge as a leading benchmark in pediatric healthcare.

\newpage

{
    \small
    \bibliographystyle{ieeenat_fullname}
    \bibliography{main}

@String(CVPR= {IEEE Conf. Comput. Vis. Pattern Recog.})

@String(CVPR  = {CVPR})

@article{armand2016gait,
  title={Gait analysis in children with cerebral palsy},
  author={Armand, St{\'e}phane and Decoulon, Geraldo and Bonnefoy-Mazure, Alice},
  journal={EFORT open reviews},
  volume={1},
  number={12},
  pages={448--460},
  year={2016},
  publisher={The British Editorial Society of Bone and Joint Surgery}
}

@article{tan2026gaitdynamics,
  title={GaitDynamics: A generative foundation model for analyzing human walking and running},
  author={Tan, Tian and Van Wouwe, Tom and Werling, Keenon F and Liu, C Karen and Delp, Scott L and Hicks, Jennifer L and Chaudhari, Akshay S},
  journal={Nature Biomedical Engineering},
  pages={1--13},
  year={2026},
  publisher={Nature Publishing Group UK London}
}

@article{read2003edinburgh,
  title={Edinburgh visual gait score for use in cerebral palsy},
  author={Read, Heather S and Hazlewood, M Elizabeth and Hillman, Susan J and Prescott, Robin J and Robb, James E},
  journal={Journal of pediatric orthopaedics},
  volume={23},
  number={3},
  pages={296--301},
  year={2003},
  publisher={LWW}
}

@article{maathuis2005gait,
  title={Gait in children with cerebral palsy: observer reliability of Physician Rating Scale and Edinburgh Visual Gait Analysis Interval Testing scale},
  author={Maathuis, Karel GB and Van Der Schans, Cees P and Van Iperen, Andries and Rietman, Hans S and Geertzen, Jan HB},
  journal={Journal of Pediatric Orthopaedics},
  volume={25},
  number={3},
  pages={268--272},
  year={2005},
  publisher={LWW}
}

@inproceedings{zhu2023motionbert,
  title={Motionbert: A unified perspective on learning human motion representations},
  author={Zhu, Wentao and Ma, Xiaoxuan and Liu, Zhaoyang and Liu, Libin and Wu, Wayne and Wang, Yizhou},
  booktitle={Proceedings of the IEEE/CVF international conference on computer vision},
  pages={15085--15099},
  year={2023}
}

@article{vielemeyer2026full,
  title={A Full-Body Motion Capture Gait Dataset of Healthy Young Adults Walking Ramps Up and Down},
  author={Vielemeyer, Johanna and Tronicke, Lisa and Schreff, Lucas and Abel, Rainer and Lechler, Knut and M{\"u}ller, Roy},
  journal={Scientific Data},
  year={2026},
  publisher={Nature Publishing Group UK London}
}

@article{sutherland1997development,
  title={The development of mature gait},
  author={Sutherland, D},
  journal={Gait \& posture},
  volume={6},
  number={2},
  pages={163--170},
  year={1997},
  publisher={Elsevier}
}

@article{carion2025sam,
  title={Sam 3: Segment anything with concepts},
  author={Carion, Nicolas and Gustafson, Laura and Hu, Yuan-Ting and Debnath, Shoubhik and Hu, Ronghang and Suris, Didac and Ryali, Chaitanya and Alwala, Kalyan Vasudev and Khedr, Haitham and Huang, Andrew and others},
  journal={arXiv preprint arXiv:2511.16719},
  year={2025}
}

@inproceedings{khirodkar2024sapiens,
  title={Sapiens: Foundation for human vision models},
  author={Khirodkar, Rawal and Bagautdinov, Timur and Martinez, Julieta and Zhaoen, Su and James, Austin and Selednik, Peter and Anderson, Stuart and Saito, Shunsuke},
  booktitle={European Conference on Computer Vision},
  pages={206--228},
  year={2024},
  organization={Springer}
}

@inproceedings{fan2023opengait,
  title={Opengait: Revisiting gait recognition towards better practicality},
  author={Fan, Chao and Liang, Junhao and Shen, Chuanfu and Hou, Saihui and Huang, Yongzhen and Yu, Shiqi},
  booktitle={Proceedings of the IEEE/CVF conference on computer vision and pattern recognition},
  pages={9707--9716},
  year={2023}
}

@article{rathinam2014observational,
  title={Observational gait assessment tools in paediatrics--a systematic review},
  author={Rathinam, Chandrasekar and Bateman, Andrew and Peirson, Janet and Skinner, Jane},
  journal={Gait \& posture},
  volume={40},
  number={2},
  pages={279--285},
  year={2014},
  publisher={Elsevier}
}

@article{baker2006gait,
  title={Gait analysis methods in rehabilitation},
  author={Baker, Richard},
  journal={Journal of neuroengineering and rehabilitation},
  volume={3},
  number={1},
  pages={4},
  year={2006},
  publisher={Springer}
}

@article{dickens2006validation,
  title={Validation of a visual gait assessment scale for children with hemiplegic cerebral palsy},
  author={Dickens, Wendy E and Smith, Michael F},
  journal={Gait \& posture},
  volume={23},
  number={1},
  pages={78--82},
  year={2006},
  publisher={Elsevier}
}

@article{deluca1997alterations,
  title={Alterations in surgical decision making in patients with cerebral palsy based on three-dimensional gait analysis},
  author={DeLuca, Peter A and Davis, Roy B and {\~O}unpuu, Sylvia and Rose, Sally and Sirkin, Robert},
  journal={Journal of Pediatric Orthopaedics},
  volume={17},
  number={5},
  pages={608--614},
  year={1997},
  publisher={LWW}
}

@article{cook2003gait,
  title={Gait analysis alters decision-making in cerebral palsy},
  author={Cook, Robert E and Schneider, Ingo and Hazlewood, M Elizabeth and Hillman, Susan J and Robb, James E},
  journal={Journal of pediatric orthopaedics},
  volume={23},
  number={3},
  pages={292--295},
  year={2003},
  publisher={LWW}
}

@article{baker2016gait,
  title={Gait analysis: clinical facts},
  author={Baker, Richard and Esquenazi, Alberto and Benedetti, Maria Grazia and Desloovere, Kaat and others},
  journal={Eur. J. Phys. Rehabil. Med},
  volume={52},
  number={4},
  pages={560--574},
  year={2016}
}

@article{bonanno2023gait,
  title={Gait analysis in neurorehabilitation: from research to clinical practice},
  author={Bonanno, Mirjam and De Nunzio, Alessandro Marco and Quartarone, Angelo and Militi, Annalisa and Petralito, Francesco and Calabr{\`o}, Rocco Salvatore},
  journal={Bioengineering},
  volume={10},
  number={7},
  pages={785},
  year={2023},
  publisher={MDPI}
}

@article{sharma2024factors,
  title={Factors influencing the clinical adoption of quantitative gait analysis technology with a focus on clinical efficacy and clinician perspectives: A scoping review},
  author={Sharma, Yashoda and Cheung, Lovisa and Patterson, Kara K and Iaboni, Andrea},
  journal={Gait \& Posture},
  volume={108},
  pages={228--242},
  year={2024},
  publisher={Elsevier}
}

@article{ben2023quantitative,
  title={Quantitative gait analysis and prediction using artificial intelligence for patients with gait disorders},
  author={Ben Chaabane, Nawel and Conze, Pierre-Henri and Lempereur, Mathieu and Quellec, Gwenol{\'e} and R{\'e}my-N{\'e}ris, Olivier and Brochard, Sylvain and Cochener, B{\'e}atrice and Lamard, Mathieu},
  journal={Scientific Reports},
  volume={13},
  number={1},
  pages={23099},
  year={2023},
  publisher={Nature Publishing Group UK London}
}

@article{ranjan2025computer,
  title={Computer vision for clinical gait analysis: A gait abnormality video dataset},
  author={Ranjan, Rahm and Ahmedt-Aristizabal, David and Armin, Mohammad Ali and Kim, Juno},
  journal={IEEE Access},
  volume={13},
  pages={45321--45339},
  year={2025},
  publisher={IEEE}
}

@article{chen2022computer,
  title={Computer vision and machine learning-based gait pattern recognition for flat fall prediction},
  author={Chen, Biao and Chen, Chaoyang and Hu, Jie and Sayeed, Zain and Qi, Jin and Darwiche, Hussein F and Little, Bryan E and Lou, Shenna and Darwish, Muhammad and Foote, Christopher and others},
  journal={Sensors},
  volume={22},
  number={20},
  pages={7960},
  year={2022},
  publisher={MDPI}
}

@article{dobson2007gait,
  title={Gait classification in children with cerebral palsy: a systematic review},
  author={Dobson, Fiona and Morris, Meg E and Baker, Richard and Graham, H Kerr},
  journal={Gait \& posture},
  volume={25},
  number={1},
  pages={140--152},
  year={2007},
  publisher={Elsevier}
}

@misc{li2026kaggle,
    author = {Li, Boyi and Shen, Yifan and Yang, Houze and Cao, Xu and Yun, Guojun and Gao, Li and Chen, Turong and Xu, Long and Cao, Jianguo and Huang, Meihuan},
    title = {[CVPR 2026] The First AI for Children Challenge},
    year = {2026},
    howpublished = {\url{https://kaggle.com/competitions/cvpr-2026-the-first-ai-children-challenge}},
    note = {Kaggle}
}

@article{yun2023selective,
  title={Selective motor control correlates with gross motor ability, functional balance and gait performance in ambulant children with bilateral spastic cerebral palsy},
  author={Yun, Guojun and Huang, Meihuan and Cao, Jianguo and Hu, Xianming},
  journal={Gait \& Posture},
  volume={99},
  pages={9--13},
  year={2023},
  publisher={Elsevier}
}
}


\end{document}